# Design and Fabrication of String-driven Origami Robots

Peiwen Yang[1], Shuguang Li[1*]

*Abstract*— Origami designs and structures have been widely used in many fields, such as morphing structures, robotics, and metamaterials. However, the design and fabrication of origami structures rely on human experiences and skills, which are both time and labor-consuming. In this paper, we present a rapid design and fabrication method for string-driven origami structures and robots. We developed an origami design software to generate desired crease patterns based on analytical models and Evolution Strategies (ES). Additionally, the software can automatically produce 3D models of origami designs. We then used a dual-material 3D printer to fabricate those wrapping-based origami structures with the required mechanical properties. We utilized Twisted String Actuators (TSAs) to fold the target 3D structures from flat plates. To demonstrate the capability of these techniques, we built and tested an origami crawling robot and an origami robotic arm using 3D-printed origami structures driven by TSAs.

## I. INTRODUCTION

Origami is a process of folding flat sheets, panels, or special laminated plates in a certain way to form a 3D structure. These origami structures have been widely used in engineering thanks to their attractive properties such as high strength, ultra-lightweight, and foldability. Applications include making static structures [1], foldable clothes, origami stents [2], origami robots [3], origami-based grippers [4], solar panels [5], and metamaterials [6].

However, the design and fabrication of complex origami structures still mainly rely on human experiences and skills, which is both time and labor-consuming. For example, when designing a waterbomb crease pattern on an A4 paper, the size of basic units depends on the density of the tessellation. If the density is changed according to specific needs, we need to manually modify the length and angle of the creases to form basic units with new sizes. Additionally, it takes two hours or longer to fold an A4-size waterbomb with a tessellation density of 7×9 based on our empirical tests. Indeed, origami researchers can enhance the ease of folding origami structures significantly through techniques such as softening creases, increasing the stiffness of origami panels, and fabricating laminated structures that facilitate self-folding [7]. However, this comes with increased difficulty and complexity in material preparation and component assembly.

To address these issues, we propose a rapid design and fabrication strategy for string-driven origami structures. The first part is a software called PyGamic that can design origami crease patterns using analytical models and optimization algorithms. It can also automatically generate 3D-printable models of origami structures for dual-material 3D printers. The second part is a device to fold origami structures with Twisted String Actuators (TSAs) [8]. In section IV (C), we demonstrate an origami robotic arm and an origami crawling robot that are both driven by onboard TSAs. The main contributions of this paper are summarized as follows:

(1) A software that can rapidly design origami crease patterns and automatically generate 3D-printable models of origami structures;

(2) A novel strategy and preliminary devices for folding origami structures using TSAs;

(3) A series of validation tests and demonstrations for the proposed rapid design and fabrication method.

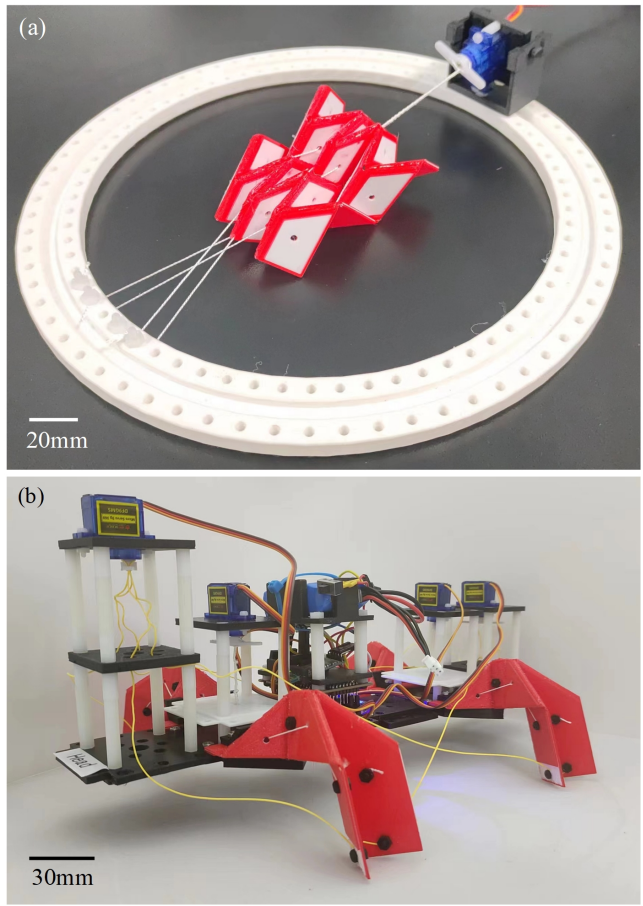

Fig. 1. Demonstration of string-driven origami structures. (a) A Miura structure folded by a TSA on a ring-shaped device. (b) A movable crawling robot with origami legs driven by onboard TSAs.

---

[1]Peiwen Yang and Shuguang Li are with the Department of Mechanical Engineering at Tsinghua University, Beijing, 100084, China.

[*]Corresponding author: Shuguang Li (e-mail: lisglab@tsinghua.edu.cn).

## II. RELATED WORKS

### A. Rapid design of origami structures

Several excellent algorithms and software tools have previously been developed for designing origami structures. Demaine and Tachi [1] developed a software called Origamizer with an analytical algorithm to generate origami crease patterns for folding any polyhedron. Dudte et al. [9] proposed a method for designing Miura patterns according to different Gaussian curvatures and unit densities. Zimmermann et al. [10] presented a design algorithm for rigid origami. With the Principle of Three Units [11], the algorithm can generate crease patterns by continuously extending three output creases from a particular node. Jin et al. [12] created an algorithm for designing the origami that matches a 3D curved surface. In general, most of the existing studies focus on rapidly designing complex crease patterns for thin sheets. Few studies in this field consider designing structures of origami panels automatically and making the fabrication process of origami structures easier.

### B. Self-folding of origami structures

In order to fold origami structures automatically, various self-folding strategies have been proposed at different scales. For small-scale origami, thermo-responsive materials or shape memory composites were widely used at the creases or joints to drive the folding process [3, 7, 13-17]. In addition, rotary motors were used to drive large-scale origami panels and structures [18, 19]. Fluid pressure has also been utilized to actuate the folding or unfolding process of origami structures with different sizes [4, 20, 21]. Due to its simplicity of construction and actuation, cable or string-driven self-folding strategy has also been widely studied for large-scale origami structures. Kilian et al. [22] used strings to fold origami with curved creases. Yim et al. [23] studied the additive folding of animatronic soft robots driven by strings. Pehrson et al. [24] presented a type of cable-driven origami-based array for spacecraft. Niu et al. [25] investigated folding origami structures via pull-up nets. Wang et al. [26] placed a string-winding motor in the center of the origami parabolic reflector to fold the whole structure.

For the cable-driven self-folding techniques, existing studies mainly focus on the self-folding of origami with serial mechanisms [23, 25] or center-symmetric structures [22, 24, 26]. Few previous studies explored the self-folding strategy for more general origami structures.

## III. ORIGAMI DESIGN SOFTWARE

### A. Design of origami tessellation

We developed an origami design software called PyGamic that can help design origami tessellation patterns. We propose a model called Transition Graph (TG) to describe the Miura-based tessellation patterns (Miura-ori). As shown in Fig. 2a & 2b, a TG consists of $\mathbb{R}^2$ continuous line segments with different lengths ($l_i$) and absolute angles ($\alpha_i$). A transition start point ($T_s$) and a transition endpoint ($T_e$) are in TG to show the transition direction. Every line segment has an entry flag ($EF_i$) indicating the type of the crease parallel to the line segment (We call it the main crease, and $\theta$ is its folding angle). The value 1 (or symbol $\vee$) represents a valley crease, and 0 (or symbol $\wedge$) represents a mountain crease (see the example in Fig. 2b). We further define the line segment as the transition vector ($\mathbf{v}_i = \mathbf{v}_i(l_i, \alpha_i, EF_i)$). Note that any two adjacent transition vectors have different entry flags.

Then, we can design the shape angle ($\beta_i$) of the Miura pattern according to the total folding ratio ($p$, which is $\theta/\pi$) of the origami and the absolute angle difference between adjacent transition vectors ($\Delta\alpha_i$). The equation is as follows:

$$\beta_i = \begin{cases} \arctan \dfrac{\tan \dfrac{|\Delta\alpha_i|}{2}}{\sin \dfrac{p\pi}{2}} & ((\Delta\alpha_i \geq 0 \wedge EF_i = 0) \vee (\Delta\alpha_i < 0 \wedge EF_i = 1)) \\ \pi - \arctan \dfrac{\tan \dfrac{|\Delta\alpha_i|}{2}}{\sin \dfrac{p\pi}{2}} & ((\Delta\alpha_i < 0 \wedge EF_i = 0) \vee (\Delta\alpha_i \geq 0 \wedge EF_i = 1)) \end{cases} \quad (1)$$

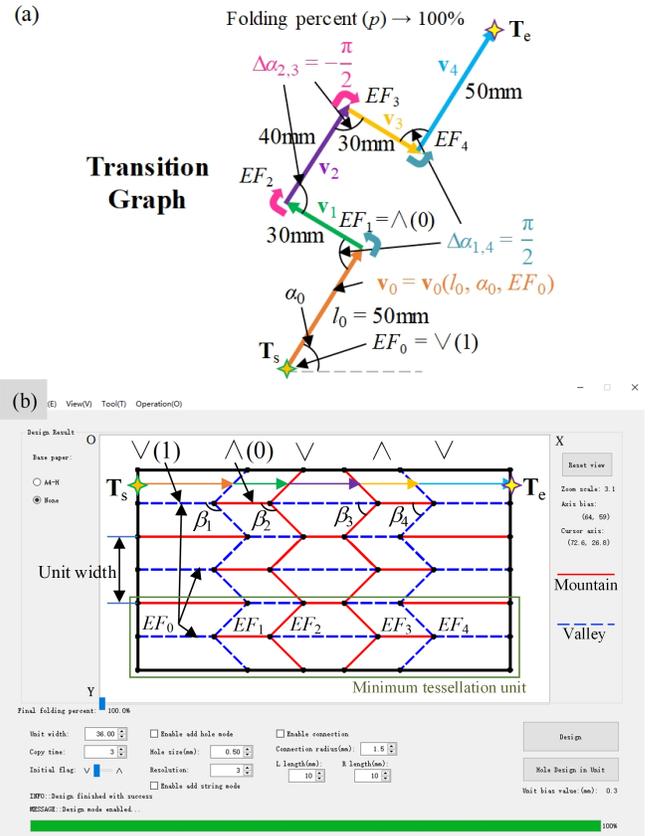

Fig. 2. An example of designing origami tessellation. (a) The user-input TG consists of five transition vectors. (b) The computed crease pattern. The unit width is 36 mm, and the copy number of the minimum tessellation unit is three. (c-e) The simulated folding process of the design at $p = 0$, 50%, and near 100%. (f-h) The real-world folding process of the origami structure.

## B. Computational design of origami robotic arms

PyGamic also enables the computational design of robotic arms with the origami-string structure [19]. The design task is shown in Fig. 3a. We first specify a $T_s$ in $\mathbb{R}^2$ space as the fixed base-link of the origami. We then mark $n_p$ ($n_p \geq 2$) $\mathbb{R}^2$ endpoints in order, requiring the $T_e$ of origami to pass through in sequence during the folding process. In Fig. 3a, we mark two points ($P_{start}$, $P_{end}$) for $T_e$ to pass through. When the origami is in an unfolded state, its transition endpoint stays at $T_{e,\,\theta=0}$, and when in a completely folded state, its transition endpoint stays at $T_{e,\,\theta=\pi}$. Additionally, we can place obstacles of any shape in $\mathbb{R}^2$ space. The movement of $T_e$ must avoid obstacles during the folding process of origami.

For every pair of adjacent transition vectors, the $\Delta\alpha_i$ can be calculated by the following equation:

$$\Delta\alpha_i = 2\arctan\left(\sin\frac{\theta}{2}\cdot\tan\beta_i\cdot(-1)^{EF_{i-1}}\right)\ (0\leq\theta\leq\pi)\quad(2)$$

Besides, we connect the initial $v_0$ to a half hypar-origami [27] unit. This ensures that the TGs of an origami string are always in $\mathbb{R}^2$ spaces parallel to each other during the folding process. Therefore, the absolute angle of the initial transition vector ($\alpha_0$) is:

$$\alpha_0 = \arccos\sqrt{\frac{2\cos\frac{\theta}{2}}{1+\cos\frac{\theta}{2}}}\ (0\leq\theta\leq\pi),\quad(3)$$

which is the variation of the transition angle of a half hypar-origami unit when it is symmetrically folded.

We employed Covariance Matrix Adaptation Evolution Strategy (CMA-ES) [28] to compute the optimal crease pattern through iterative design. By using the data $\{l_0, l_1, ..., l_n\}$, $\{\beta_1, ..., \beta_n\}$ and $EF_0$ generated by the algorithm, for any given $\theta$, PyGamic calculates every $v_i$ according to equations (2) and (3). Note that $n$ refers to the number of Miura-ori in one minimum tessellation unit (see Fig. 2b). Then $T_e$ is calculated by:

$$T_e = T_s + \sum_{i=0}^{n} v_i \quad(4)$$

PyGamic samples $T_e$ with the increase $\theta$-step of 4° from $\theta = 0$ to $\pi$, and gets 46 transition endpoints. In the evaluation step of the algorithm, we define the fitness function as follows:

$$f = \left(\frac{\alpha}{10+\|T_{e,\,\theta=0}-P_{start}\|} + \frac{(600-\alpha)}{10+\|T_{e,\,\theta=\pi}-P_{end}\|}\right) - 4N,\quad(5)$$

where $\alpha$ is a tunable hyperparameter controlling the reward to $T_e$ for matching the target points; $N$ means the number of improper states within 46 sampled states, such as self-intersection of transition vectors and moving $T_e$ into the warning areas. In the example shown in Fig. 3, we set $\alpha = 120$ to encourage $T_{e,\,\theta=\pi}$ to be closer to $P_{end}$.

PyGamic runs ten independent evolution processes for every design task and then outputs the top-ranking design results. Each result includes an origami crease pattern and the moving trajectory of $T_e$ (as shown in Fig. 3b & 3c).

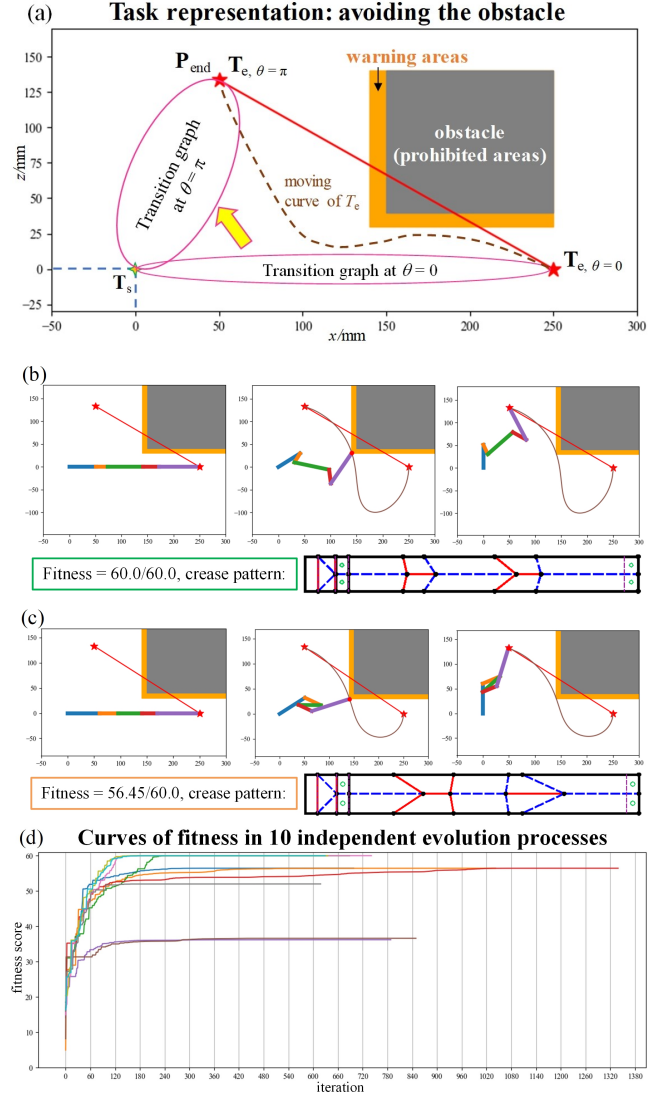

Fig. 3. A computational design example of the origami robotic arm. (a) Defining the computational design task: $T_s$ of the origami was fixed at (0, 0), and $P_{start}$ (250, 0) and $P_{end}$ (50, 133.3) were marked. Then, we introduced warning areas ($x \geq 140 \wedge y \geq 30$) and prohibited areas ($x \geq 150 \wedge y \geq 40$). (b, c) Some top-ranking results generated by PyGamic. (d) The evolution processes of designs. The fitness of the optimal solution improved with increased iterations. The final fitness scores were not the same because of the different initial mean and variance of CMA-ES.

## C. Automatically generating models of origami structures

After the design process, PyGamic can automatically generate 3D models of origami design results, which form specific origami structures. For a designed crease pattern, the user can define the height and inner bias of origami panels. As shown in Fig. 4a, PyGamic can then automatically calculate the maximum operation region of each panel (orange dash-dotted lines), where holes (green circles) can be automatically or manually added. After the structure design is completed, PyGamic calculates the triangular mesh of panels and creases and output four Stereolithography (STL) files, as shown in Fig. 4b. These files can then be assembled into a specific origami structure (See details in section IV).

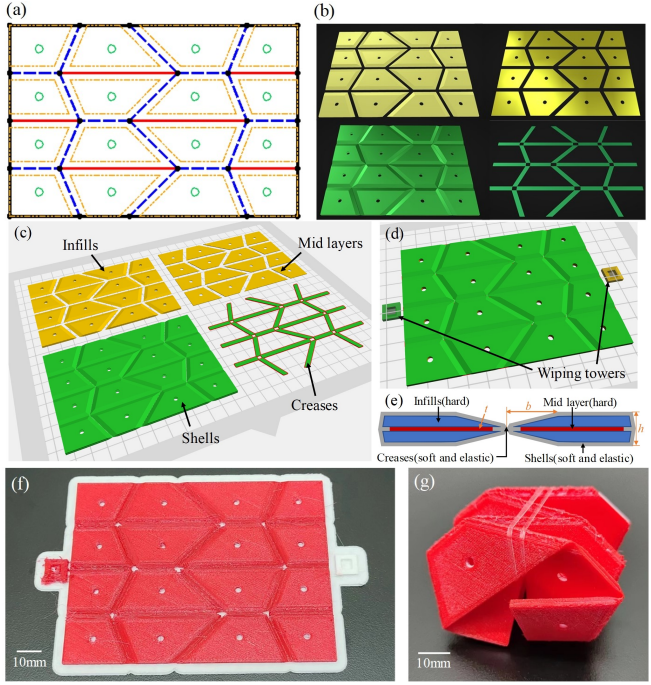

Fig. 4. Design and 3D-printing of origami structures. (a) Designed panels with specific size and hole structures. Holes were automatically added to the center of each panel. (b) The four STL files were generated by PyGamic. They are infills (top-left), mid-layers (top-right), shells (bottom-left), and creases (bottom-right). (c) The imported four STL files are shown in the slicing software. (d) The assembly of the origami structure parts. (e) The cross-section of the origami panel. (f) The printed origami structure. The materials used are PLA (white) and TPU95 (red). (g) The folded state of the origami structure.

## IV. FABRICATION OF ORIGAMI STRUCTURES

### A. Manufacturing origami structures by 3D printing

We used a dual-material 3D printer (Flashforge Creator Pro 2) to print a wrapping-based origami structure with excellent mechanical properties [29]. As shown in Fig. 4c, four generated STL files were imported into the slicing software (Flashprint 5). We assembled these parts in the slicing software to form a complete printable object (Fig. 4d). A cross-sectional view of the origami structure perpendicular to the crease is shown in Fig. 4e. The material of the shells and the creases (shown in green) is Thermoplastic Polyurethane (TPU), which is soft and elastic. In contrast, the infills and the mid-layers (shown in yellow) are made of Polylactic Acid (PLA), which is a hard material. Every part has its own functions:

- Hard infills enhance the bending strength of the origami panels;
- Soft and elastic creases have high fatigue strength and can be folded repeatedly;
- Soft shells connect the creases to each other, ensuring that the printed creases do not detach from the panels;
- Hard mid-layers connect the upper and lower parts of the infills. The inner bias of mid-layers is slightly larger than the infills to leave more space for printing creases, improving the success rate of printing.

If the inner bias of every origami panel is $b$, the maximum height of the shell structure is $h$, and the thickness of mid-layers and creases is $t$, then we can calculate the maximum folding angle of the origami structure using:

$$\theta_{\max} = 2\arctan\frac{2b}{h-t} \qquad (6)$$

In the example shown in Fig. 4f, the values for the parameters were $b = 3$ mm, $h = 2.2$ mm, and $t = 0.4$ mm. The calculated $\theta_{\max} \approx 146.6°$, and $p_{\max} \approx 81.4\%$.

It should be noted that the calculation of the maximum folding angle mentioned in equation (6) is applicable to creases without multi-layer thick-panel constraints. When there is some single vertex with multiple creases in the origami structure, some creases may be severely stretched in the width direction due to multi-layer thick-panel constraints. In this case, the maximum folding angle of these creases cannot be calculated using the above equation.

### B. Folding origami structures by external TSAs

Once the flat origami is printed, the next step is to fold it. A self-folding method based on TSA was introduced. As shown in Fig. 5d, multiple TSAs can be placed on a ring-shaped slide rail. To fold the origami structure with TSAs, we place it on the center of the slide rail and route strings through the holes in the origami structure following a designed routing path.

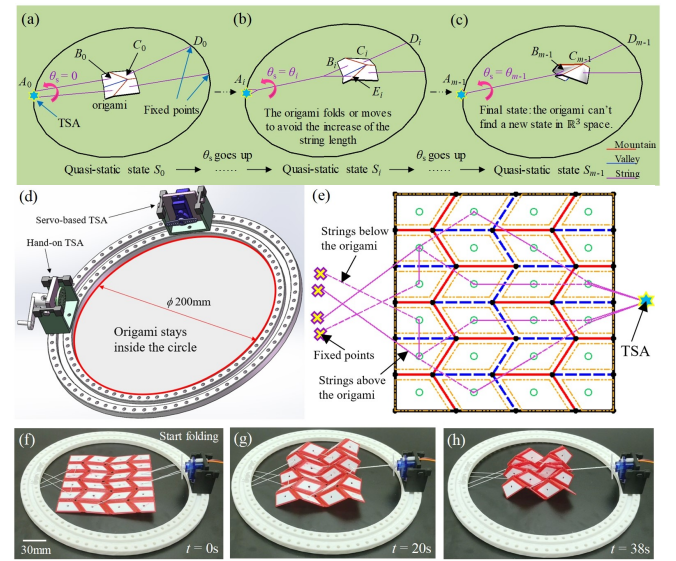

Fig. 5. An example of folding origami structures using external TSAs. (a-c) The explanation of the DCQC principle. From the initial state (a) to the final state (c), the Miura-ori folds itself to satisfy the displacement constraints. In this case, $l_{A_iB_i}$ increases at the cost of reducing $l_{B_iC_i}$ and $l_{C_iD_i}$. $B_i$, $C_i$, and $D_i$ tend to be collinear. (d) The designed ring-shaped slide rail with an inner circular space. (e) A string routing path designed in PyGamic according to the DCQC principle. We used a TSA with four strings to fold a Miura-ori structure. (f-h) The folding process of the printed Miura-ori structure. To better see the panels and creases, we did not print the upper cover of the soft shell in this demonstration.

We established a principle called Displacement Constraints under Quasi-static Conditions (DCQC) to help us design feasible methods of routing strings among holes. Fig. 5a-5c shows a simulation of folding a Miura-ori using a TSA. Assuming that the non-stretchable strings are tight in the initial state, the Miura-ori is placed at the center of the circle, and all the holes are at the center of each origami panel. When the system is in quasi-static moment $S_0$, the initial length of a string in TSA is determined by:

$$L_0 = l_{A_0B_0} + l_{B_0C_0} + l_{C_0D_0} = A_0B_0 + B_0C_0 + C_0D_0, \quad (7)$$

where $l_{A_0B_0}$ refers to the remaining string length between $A_0$ and $B_0$, and $A_0B_0$ is the distance between $A_0$ and $B_0$. Here, $l_{A_0B_0}$ equals $A_0B_0$ as the string is tight (similarly, $l_{B_0C_0} = B_0C_0$, and $l_{C_0D_0} = C_0D_0$). Let $\theta_s$ be the rotation angle of the TSA, for any quasi-static moment $S_i$ with $\theta_s = \theta_i$ (Fig. 5b), the new remaining string length between $A_i$ and $B_i$ is calculated by:

$$l_{A_iB_i} = \begin{cases} \sqrt{\|\mathbf{x}_{dis}\|^2 + \frac{(d_2 - d_1\cos\theta_i)^2 + (d_1\sin\theta_i)^2}{4}}, & 0 \leq \theta_i < \pi \\ \sqrt{\|\mathbf{x}_{dis}\|^2 + \frac{(d_2 + d_1 + (\theta_i - \pi)d_s)^2}{4}}, & \theta_i \geq \pi \end{cases}, \quad (8)$$

where $\|\mathbf{x}_{dis}\|$ is the distance between the rotation center of the TSA and the middle point of two first-entry holes of strings (in Fig. 5b, it is the middle point of $B_iE_i$), $d_1$ is the rotation diameter of the TSA, $d_2$ is the distance between two first-entry holes of strings (in Fig. 5b, it is $B_iE_i$), and $d_s$ is the width of the string. Note that $l_{A_iB_i}(\theta_s = 0) = A_0B_0$.

Then we consider any two adjacent quasi-static moments $S_{i-1}$ and $S_i$ ($i = 1, 2, ..., m$-1; note that $m$ indicates the total number of quasi-static states). Starting from the moment $S_{i-1}$, the TSA needs to rotate $\theta_i - \theta_{i-1}$ more to reach the moment $S_i$. This will result in an increase of $l_{A_iB_i}$ if the origami is not moved. Then, the total length of the string will increase, thereby exceeding $L_0$. The solution to this case is that the origami needs to be folded or moved, changing the values of $l_{A_0B_0}$, $l_{B_0C_0}$, and $l_{C_0D_0}$ to update the string length back to $L_0$. If the solution exists, the system can successfully reach $S_i$. Otherwise, $S_{i-1}$ is the final state.

Therefore, a proper string routing method should enable the origami folding process to satisfy the DCQC continuously. For example, considering a fold-in-half unit with a mountain (valley) crease, the string between the holes in the two adjacent panels should be below (above) the crease.

Following the DCQC principle, we planned a string routing path to help fold a $120 \times 120$ mm$^2$ origami with 6×4 Miura tessellation (Fig. 5e). The real-world folding process is shown in Fig. 5f-5h.

### C. Origami robots driven by onboard TSAs

Besides folding origami structures with external TSAs, we can install TSAs on the origami structure. We developed a driver module (weight: 109.71 g) containing a battery and a controller, combined with TSA modules and origami structures to build origami robots. First, we made a boat-like origami structure (weight: 85.66 g) shown in Fig. 6. The crease pattern (Fig. 6a), with a size of $165.6 \times 96.6$ mm$^2$, was defined in a Drawing Exchange Format (DXF) file, which was imported into PyGamic. After designing structures of holes and panels, PyGamic exported four STL files for 3D printing. We installed a TSA module on the printed boat-like origami structure (Fig. 6b & 6c). After plugging in the driver module, the TSA produced tension on the strings and pulled the origami structure to fold up (Fig. 6d).

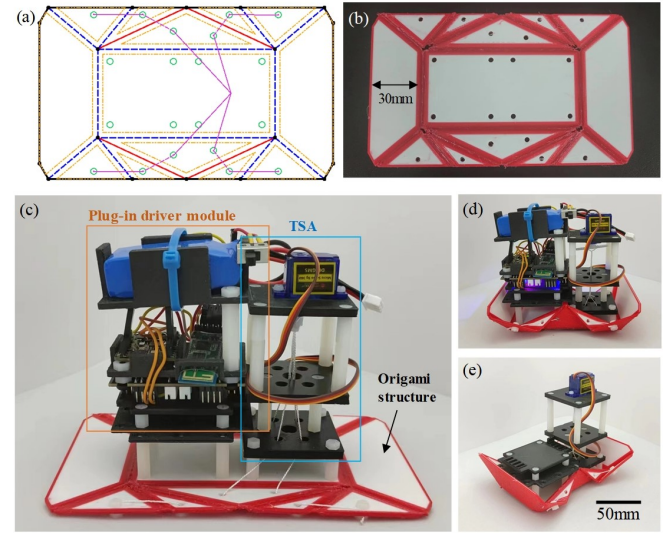

Fig. 6. Design and fabrication of the boat-like origami structure. (a) The designed crease pattern, structure of panels, location of holes, and string routing path (shown in purple) in PyGamic. (b) The printed origami structure. (c) The assembled boat-like origami structure. (d) The folded state of the origami structure. (e) The origami structure could remain folded when the driver module was removed.

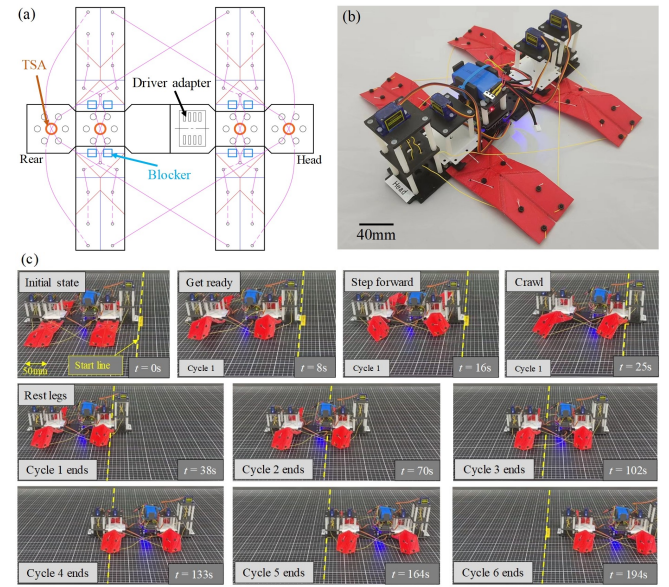

Fig. 7. Design, fabrication, and control of the origami crawling robot. (a) The designed drawing of the origami robot. Every TSA rotates four strings. The blockers limit the upper bound of the folding angle of creases covered by blockers (up to 60°). The driver adapter contains pins of all servos, providing a convenient way to connect to the driver module. (b) The assembled origami robot in the real world. (c) The motion of the robot. A movement cycle includes getting ready, stepping forward, crawling, and relaxing legs. The robot crawled forward 380 mm (one body length) in 194 seconds.

We further designed and built a crawling robot (weight: 361.45 g) with four origami legs (Fig. 7a & 7b). We used four TSA modules to control the robot. The size of the robot is 340 × 320 × 122 mm$^3$. The cycle of a crawling motion was about 32 seconds. The robot moved forward 63 mm on average for every crawling cycle on a PVC cutting mat.

Finally, we built an origami robotic arm (weight: 108.00 g). The designed origami crease pattern is shown in Fig. 3b, which is the best solution in the given computational design task. We replicated the scene in the previous simulation, as shown in Fig. 8b. We routed strings parallel to transition vectors, and used a TSA to fold the origami robotic arm. It managed to avoid the obstacle (Fig. 8h) and reach the toy frog (Fig. 8j).

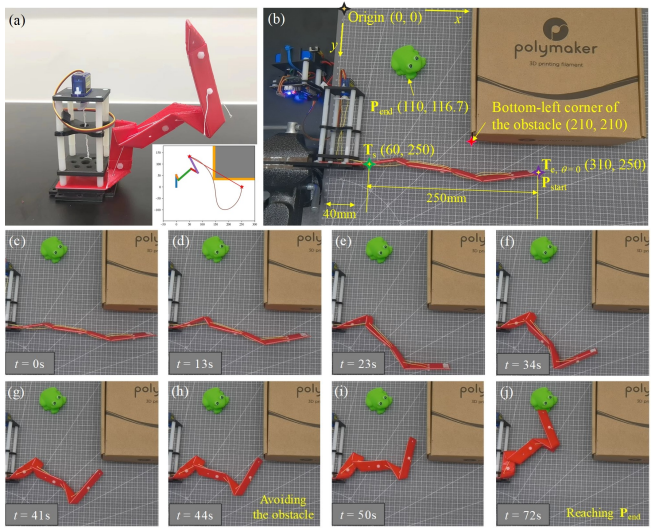

Fig. 8. Design, fabrication, and control of the origami robotic arm. (a) The fabricated origami robotic arm. Its crease pattern comes from the solution with the highest fitness in the computational design task. (b) The working environment built for the origami robotic arm. (c-j) The motion of the arm.

## V. CONCLUSION

In this work, we presented a practical approach to designing origami crease patterns and generating their printable 3D models with dual materials. In addition, we proposed the DCQC principle to help plan the string routing path for string-driven origami structures. We successfully demonstrated the capability of the proposed method through the design, fabrication, assembly, and actuation processes of a boat-like origami structure, an origami crawling robot, and an origami robotic arm. This approach allows us to rapidly design and fabricate various origami structures or robots on demand (e.g., a lightweight origami robotic arm for a particular pick-and-place task at a low cost).

Nevertheless, some questions remain open, given the complex nature of origami structures. For example, the current string routing method and control strategy for TSAs are empirically driven and may need to be optimized. Additionally, different physical parameters, such as different output torques of TSAs, different stiffnesses of creases, or different densities of origami panels, may result in different string routing designs for origami structures with the same morphology. Therefore, in our future work, we plan to consider the system's physical characterizations and then develop proper AI algorithms to design optimized string routing paths and TSA control strategies.

ACKNOWLEDGMENT

This work was supported in part by the Tsinghua University Initiative Scientific Research Program (2022Z11QYJ002). We thank Shu Leng, Paul Beckers (Tsinghua University) and Haoruo Zhang (The University of California, San Diego) for proofreading the manuscript.